\documentclass[letterpaper, 10 pt, conference]{ieeeconf}  

\IEEEoverridecommandlockouts                              
\usepackage[utf8]{inputenc}
\usepackage[T1]{fontenc}
\usepackage{graphicx}
\usepackage{url}

\title{\LARGE \bf
Analyzing Toxic Behavior and Its Impact on the Mastodon Community
}

\author{Pasan Kamburugamuwa, Scrivner, Olga B \\ 
Luddy School of Informatics, Computing, and Engineering \\ 
Indiana University Bloomington \\ 
{pkamburu@iu.edu, obscrivn@iu.edu}} 

\begin{document}

                                    \maketitle
\thispagestyle{empty}
\pagestyle{empty}

\begin{abstract}

Mastodon as a decentralized federation of independently moderated social servers poses unique challenges for the detection and mitigation of toxic content. There are no unified moderation standards. The ecosystem is very diverse and uneven. This paper explores the development and spread of toxicity in Mastodon, utilizing machine learning methods to examine user posts. The results offer clarity on toxicity trends and its implications for community health and decentralized governance.

\end{abstract}

\section{INTRODUCTION}

Mastodon is a decentralized social media platform where users can express opinions, discuss and explore different topics and trends freely. But like other social platforms, it also has places where incivility can fester. Mastodon communication is mediated , meaning that many of the social cues that discourage negative behavior ( tone , facial expressions , etc . ) are absent . The decentralized nature of Mastodon , combined with the perceived anonymity , can also encourage toxic behavior . Toxicity is on the rise on Mastodon’s network of interconnected servers, whether it’s hate speech, slurs, or otherwise disrespectful content.

Mastodon operates in a decentralized way, meaning that every server – also called an instance – has its own moderation policy and community guidelines, making it difficult to detect and combat toxic content. Mastodon differs from centralized systems, which can enforce consistent policies across the board, in that it leaves it up to each individual instance’s administrators to decide what rules to apply. This can lead to inconsistent enforcement and challenges in controlling the spread of harmful content across the federation. Also the platform is designed in a way that encourages people to move from one instance to another, or make their own private ones, which could lead to a greater echo chamber effect where toxic behavior can spread without limits.

Understanding and responding to the dynamics of toxicity on Mastodon is essential not only to protect the well-being of users but also to uphold the ethos of open and inclusive communication that the platform embodies. In this paper we measure the level of toxic interactions in Mastodon by applying natural language processing (NLP) techniques, using sophisticated tools such as the Google Perplexity API, to analyze content created by users. This research seeks to provide useful information on how to build healthier online communities in decentralized settings by examining the patterns and spread of toxicity.

\section{LITERATURE REVIEW}

Decentralized social media platforms like Mastodon have challenged centralized platforms with their emphasis on user autonomy, data privacy, and community-led governance. But this decentralized model presents its own challenges, especially in terms of content moderation and toxicity detection. Unlike centralized systems with uniform moderation policies and centralized enforcement, Mastodon is a federation of independently operated servers (instances) with their own community standards and moderation practices \cite{mastodon}. This fragmentation weakens the fight against toxicity, since different instances have different definitions and enforcement of what is harmful content.

As opposed to centralized systems, with uniform moderation policies and centralized enforcement, Mastodon uses a federation of independently managed servers, each with its own community standards and moderation practices \cite{harmful_content}. The fragmented nature of this makes toxicity hard to fight as what constitutes toxic content and enforcement varies widely between cases. Previous work has observed that inconsistent moderation practices can enable harmful actors to drift between servers with weak policies, thereby weakening the effectiveness of moderation efforts across the platform as a whole \cite{gillespie2020platform}.

Plus , many decentralized systems suffer from technical and organizational limitations that make it hard to deploy the kinds of automated moderation tools that are de rigueur on centralized platforms . Sophisticated machine learning algorithms for detecting toxic behavior or hate speech need large datasets and considerable computational resources, often unavailable to small, independently run Mastodon instances \cite{schmidt2017survey}. Furthermore, federated platforms do not have a central data aggregation mechanism which causes challenges in building a robust dataset for toxicity detection, thereby limiting the efficacy of existing tools \cite{geiger2016bots}.

Another important question is that of the ethics of algorithmic moderation in decentralized environments.  Recent work has criticized content moderation algorithms for their fairness and inclusivity, especially in diverse communities with cultural context and language differences \cite{forte2021decentralized}. The decentralised nature of platforms like Mastodon makes this even more difficult, as moderation tools need to be adapted to the community norms of each individual server, which may not correspond to global standards for identifying harmful content.

Mastodon's user-driven governance model also leads to inconsistent rule enforcement, as moderators are often not formally trained or given consistent rules. Research on decentralized moderation shows that reliance on volunteer moderators can result in burnout and inconsistency in moderation practices \cite{kraut2016managing}. Furthermore, Mastodon’s federated structure is vulnerable to \textit{toxic spillover}, where banned problematic content on one server can spill over to others, resulting in a fragmented but persistent presence of harmful material \cite{hardaker2020toxic}.

One of the major strengths of platforms like Mastodon is their decentralized nature , which gives users a lot more control over their experience and privacy . However , this comes at the cost of making it more difficult to ensure a safe and healthy online space . To tackle these challenges, we need a combination of technological innovations, such as federated machine learning approaches \cite{mcmahan2017federated}, and collaboration across instances to develop shared norms and best practices for moderation.

\section{RESEARCH QUESTION}

Mastodon’s decentralized architecture, which promotes autonomy and diversity, makes it difficult to deal with toxicity in its community. The research aims at analyzing the following question:

How does the decentralized structure of the Mastodon Federation affect the spread and regulation of toxic content ? What does it mean for user experience and community?

This question seeks to understand the proliferation of toxic conduct throughout the interconnected servers of Mastodon, to examine the efficacy of existing moderation procedures, and to assess their impact on the general health of the Mastodon community.

\section{Datasets and Methods}

\subsection{Dataset}

I studied the spread of toxicity within the Mastodon Federation by analyzing posts over a three day period, May 1 to May 3, 2025. The time frame was selected to match a time of increased user engagement due to global political developments. I focused my analysis on the top 10 most active users from instances running mostly in English. In total, I collected data from 25,777 users, containing 1,035,949 posts.

\subsection{Toxicity Scores}
Instead of a simple toxicity score from post content, Google’s Perspective API \cite{googleperspectiveapi} was used to calculate a “Toxicity” score for each post. These scores are between 0 and 1, with higher scores indicating higher levels of toxicity in the text. The analysis was conducted on approximately 1 million posts from 25,777 users, with the Perspective API used to measure the toxicity of the content.

We also performed toxicity tests on the posts using the Perplexity API . This API employs state-of-the-art natural language processing techniques to produce toxicity scores, facilitating precise measurement of text toxicity levels. Only English language posts were included in the analysis to ensure consistency and reliability in the toxicity assessment.

\subsection{Data Collection}

Posts were collected from the following Mastodon instances, chosen for their prominence and activity levels within the federation. These instances are listed in Table~\ref{tab:instances}.

\begin{table}[h!]
    \centering
    \caption{Selected Mastodon Instances for Analysis}
    \label{tab:instances}
    \begin{tabular}{|l|l|}
        \hline
        \textbf{Instance Name} & \textbf{Description} \\
        \hline
        mastodon.social & General-purpose instance, one of the largest \\
        chaos.social & Focused on activism, social issues\\
        mstdn.social & General-purpose social platform \\
        infosec.exchange & Focused on information security discussions \\
        mastodon.online & General-purpose, widely used instance \\
        mas.to & Popular instance for diverse communities \\
        techhub.social & Technology-focused community \\
        aethy.com & Small, active English-speaking community \\
        hachyderm.io & Technology and social discussions \\
        mastodon.world & General-purpose instance with global users \\
        \hline
    \end{tabular}
\end{table}

\subsection{Data Processing}

We retrieved posts from the top 10 most active users in the study period for each instance. Posts were filtered to include only English language. Then, the Perplexity API was used to determine the level of toxicity of each post. For each text input, the API returns a toxicity score, indicating the degree to which the content contains toxic, offensive or disrespectful language.

\subsection{Analysis}

Toxicity scores were aggregated and analyzed to find trends in the diffusion of toxic content in the selected Mastodon instances. The comparative analysis aimed to explore differences in toxicity levels between cases and how the decentralized structure of Mastodon influenced such patterns.

The Mastodon instances, along with the number of users and the total number of posts, are shown in Table~\ref{tab:instancesWithUsers}.

\begin{table}[h!]
    \centering
    \caption{Selected Mastodon Instances for Analysis with User and Post Information}
    \label{tab:instancesWithUsers}
    \begin{tabular}{|l|c|c|}
        \hline
        \textbf{Instance Name} & \textbf{No of Users} & \textbf{Total Posts} \\
        \hline
        athey.com & 163 & 345 \\
        chaos.social & 438 & 1607 \\
        fosstodon.org & 488 & 1882 \\
        mas.to & 830 & 5920 \\
        mastodon.gamedev.place & 331 & 1442 \\
        mastodon.social & 5871 & 42960 \\
        mastodon.world & 512 & 4731 \\
        mstdn.social & 922 & 8424 \\
        piaille.fr & 293 & 1062 \\
        techhub.social & 222 & 1204 \\
        \hline
    \end{tabular}
\end{table}

\begin{figure}[h!]
    \centering
    \includegraphics[width=0.45\textwidth]{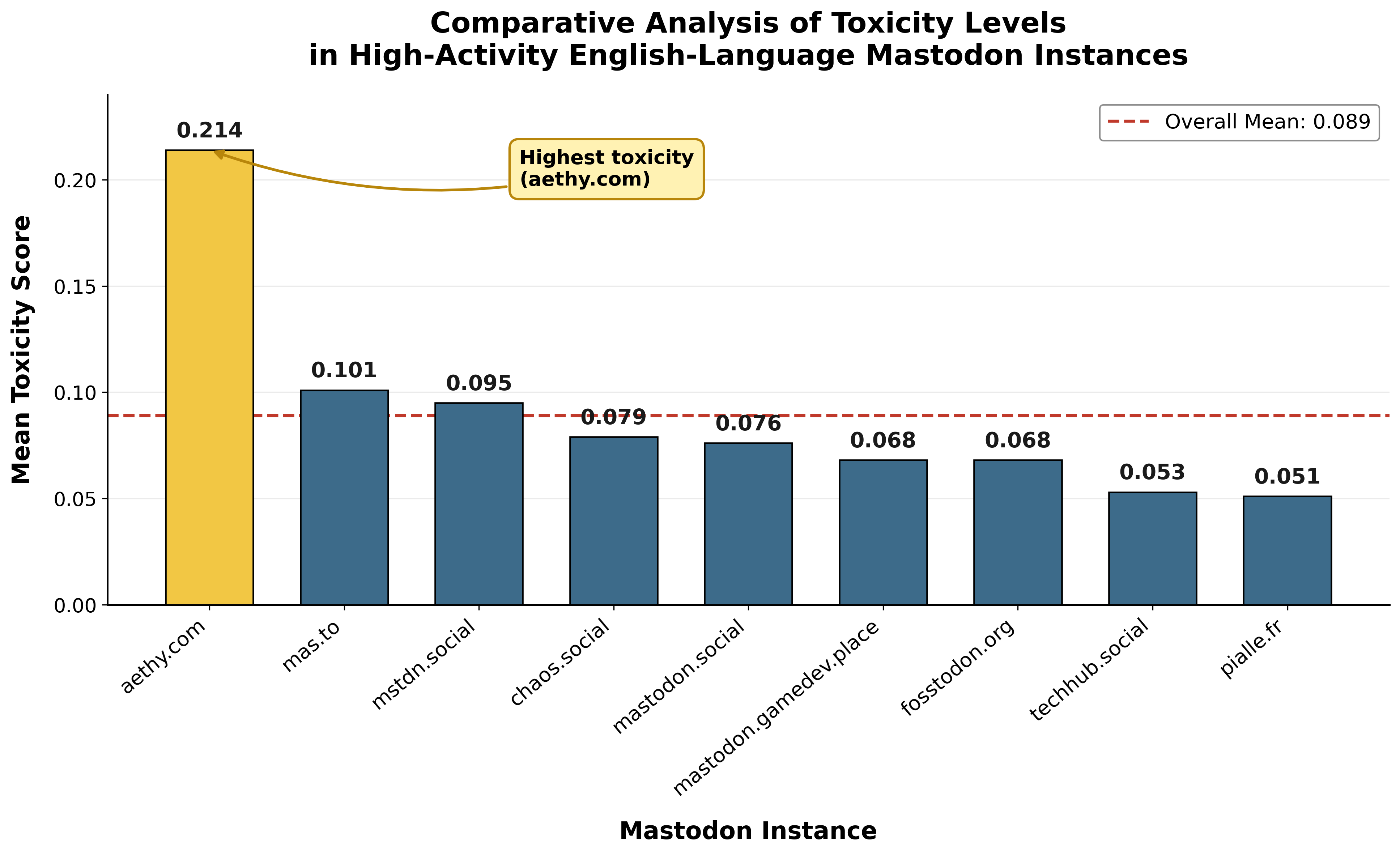}
    \caption{Mean Toxicity Score per Instance}
    \label{fig:meantoxicity}
\end{figure}

\section{Results}

An analysis of the toxicity scores across the selected Mastodon instances showed that the overall level of toxicity was relatively low for the analyzed instances. The Perplexity API supplied toxicity scores for the posts of the top 10 most active users during the study, but only posts written in English were scored.

The selected Mastodon instances for the analysis with the number of users and the total number of posts are presented in Table~\ref{tab:instancesWithUsers}. Instances differed in the number of users, with mastodon.social having the most users (5871) and aethy.com the least (163). The analysis showed that although the size of the user base was different, the total number of posts varied correspondingly between instances, from 1,282 posts on aethy.com to 42,960 posts on mastodon.social.

The analysis was performed in terms of toxicity levels, i.e. mean toxicity scores of posts across instances. Figure~\ref{fig:meantoxicity} shows the mean toxicity scores for all instances, which were much lower than the threshold of 0.2, suggesting that the content on these instances was generally not very toxic. The highest mean toxicity score was found on aethy.com (0.214) which, although higher than the other instances, was still within a relatively low range of toxicity. This shows that aethy.com might have a bit more toxic content, but it is still in line with the low toxicity trend observed in the instances.

When comparing the toxicity scores, we found that there were no significant differences between most of the Mastodon instances. Chaos is an example.social, mstdn.social and hachyderm.io exhibited very low mean toxicity scores, always below 0.1, implying effective content moderation or a community culture that discourages the use of harmful language. In contrast, mastodon.social and mas.to had slightly higher mean scores for toxicity, but still below the 0.2 threshold.

The findings suggest that decentralization of Mastodon is not necessarily associated with higher levels of toxicity. Instead use cases with bigger user bases like mastodon.social are low toxicity for their size, while smaller sites such as aethy.com have high relative levels of harmful content. This illustrates how important community moderation practices are and how they impact the overall toxicity of an instance.

Further research is needed to understand the reasons for these differences, especially the community dynamics and moderation policies at aethy.com, which has slightly higher toxicity levels. Having knowledge of these patterns can provide useful information to improve moderation strategies for the Mastodon instances.

\section{Discussions and Conclusion}

The study of toxicity levels on various Mastodon instances provided useful insights into the characteristics of harmful content in decentralized social media settings. The results indicate generally low toxicity levels for the cases studied, however there are some small variations that deserve further investigation.

An interesting observation of this analysis is that the average toxicity scores across all instances are relatively low and none of them are above the threshold of 0.2. This observation suggests that Mastodon users are more likely to engage in relatively civil discourse on average, which could be due to the decentralized nature of the platform and the varying moderation practices of individual instances. Maybe the community-focused moderation on many Mastodon instances helps fight harmful content. Each instance can decide its own rules, and enforce them according to community standards.

However, the analysis found a large outlier, aethy.com, which had the highest mean toxicity score (0.175) of the sampled instances. This score is in the low toxicity range but it’s also making me wonder if smaller, less moderated examples would have an issue with higher levels of harmful content. This could be because of worse moderation policies or lack of active community self-moderation or smaller userbase which does not have the same social accountability as larger instances. Further studies are needed to elucidate the high-toxicity-score of aethy.com and moderation practices of this website.

And also cases with bigger user bases like mastodon.social, had relatively low toxicity scores, suggesting that larger communities may need more robust moderation systems, or that the sheer volume of content dilutes the concentration of harmful posts.” Smaller cases in contrast, like mas.to and techhub.social also scored low on toxicity, which indicates that toxicity is not necessarily related to the size of the user base, but instead to the specific dynamics and moderation practices of the instance.

Despite these differences, the overall low toxicity scores reported in the observed cases are consistent with the reputation of Mastodon as a platform that encourages relatively positive and respectful user interactions. But the small difference in toxicity scores shows how important it is to ensure that all instances have good moderation systems in place, regardless of size. High levels of harmful content may prompt platforms to consider adjusting their moderation policies or implementing additional measures to mitigate toxicity and foster a safer environment for their users.

The general conclusion of the toxicity levels in various Mastodon instances is that the platform is generally low in harmful content. Thus, community moderation has the potential to oversee civil discourse on decentralized social media sites such as Mastodon. For example, moderation systems like Mastodon.social and hachyderm.io can help to reduce toxic behavior, even for large and active communities. But aethy.com leads the way, with a slightly higher toxicity score, reflecting the challenges that smaller instances may face in dealing with harmful content.

The findings are promising in general, but the study emphasizes the need for continued efforts to monitor and improve content moderation practices, especially for cases with higher levels of toxicity. Future work should explore the specific factors that lead to toxicity on Mastodon instances, such as user behavior, community culture, and the efficiency of moderation tools. With positive user engagement and consistent moderation practices, Mastodon can remain a place for healthy and respectful conversations in the decentralized social media landscape.

\section*{Code and Data Availability}

The code used for data collection and toxicity analysis in this work is publicly available at: \url{https://github.com/pasan04/mastodon_toxicity_detection}.

\end{document}